\documentclass[10pt]{article} 
\usepackage{tikz}
\usepackage{pgfplots}       
\pgfplotsset{compat=1.18}   
\usepackage{algorithm}
\usepackage{amsmath}
\usepackage{algorithm}
\usepackage{algorithmicx}
\usepackage{algpseudocode}
\usepackage{amsfonts}
\usepackage{alphalph} 

\usepackage{xcolor} 
\definecolor{mygray}{RGB}{80,80,80}
\definecolor{lightpurple}{RGB}{153,102,204}

\definecolor{YesColor}{RGB}{220,240,220}      
\definecolor{LimitedColor}{RGB}{255,245,220}  
\definecolor{NoColor}{RGB}{245,220,220}       
\definecolor{BothColor}{RGB}{220,235,255}     

\definecolor{DeepBlue}{HTML}{1f77b4}
\definecolor{BurntOrange}{HTML}{ff7f0e}
\definecolor{ForestGreen}{HTML}{2ca02c}
\definecolor{PlumPurple}{HTML}{9467bd}

\definecolor{RoyalBlue}{HTML}{0047AB}
\definecolor{CrimsonRed}{HTML}{DC143C}
\definecolor{DarkGreen}{HTML}{006400}
\definecolor{Goldenrod}{HTML}{DAA520}

\usepackage{tikz}
\usepackage{pgf-pie}
\usetikzlibrary{positioning, shapes.geometric, arrows.meta}
\usepackage{wrapfig}
\usetikzlibrary{fit}
\usepackage{subcaption} 
\tikzset{
  block/.style={
  rectangle,
  draw,
  fill=gray!10,
  thick,
  minimum width=3cm,
  minimum height=1cm,
  rounded corners,
  text width=3cm,    
  align=center       
  },
  child/.style={
  rectangle, 
  draw,
  thick,
  fill=gray!10,
  minimum width=3cm, 
  minimum height=1cm,
  rounded corners,
  text width=2.6cm, 
  align=center
  },
  arrow/.style={
  -{Stealth}, 
  thick,
  draw=gray,
  },
}

\usepackage[preprint]{tmlr}

\usepackage{enumitem}

\usepackage{amsmath,amsfonts,bm}

\def\eqref#1{equation~\ref{#1}}

\def\1{\bm{1}}

\DeclareMathAlphabet{\mathsfit}{\encodingdefault}{\sfdefault}{m}{sl}
\SetMathAlphabet{\mathsfit}{bold}{\encodingdefault}{\sfdefault}{bx}{n}

\usepackage{hyperref}
\usepackage{url}
\usepackage{booktabs}
\usepackage{adjustbox}
\usepackage{multirow}
\usepackage{mathtools}

\usepackage[normalem]{ulem}

\title{\emph{CommFuse}: Hiding Tail Latency via Communication Decomposition and Fusion for Distributed LLM Training}

\author{\name Rezaul Karim\thanks{This manuscript is an early preprint and has not been rigorously revised for typos or minor errors. Feedback is very welcome. Corresponding author: \texttt{rezaul.karim3@huawei.com}.} \email rezaul.karim3@huawei.com \\
       \addr Toronto Ascend Team
       \AND
       \name Austin Wen \email austin.wen1@huawei.com \\
       \addr Toronto Ascend Team
       \AND
       \name Wang Zongzuo \email wangzongzuo@huawei.com\\
       \addr Toronto Ascend Team 
       \AND 
       \name Weiwei Zhang \email weiwei.zhang2@huawei.com \\
       \addr Toronto Ascend Team 
       \AND
       \name Yang Liu \email yang.liu8@huawei.com \\
       \addr Toronto Ascend Team 
       \AND
       \name Walid Ahmed \email walid.ahmed1@huawei.com \\
       \addr Toronto Ascend Team              
       }

\def\month{MM}  
\def\year{YYYY} 
\def\openreview{\url{https://openreview.net/forum?id=XXXX}} 

\usepackage{hyperref}
\hypersetup{pdfborder=0 0 0}

\begin{document}

\maketitle

\begin{abstract}
The rapid growth in the size of large language models has necessitated the partitioning of computational workloads across accelerators such as GPUs, TPUs, and NPUs. However, these parallelization strategies incur substantial data communication overhead significantly hindering computational efficiency. While communication-computation overlap presents a promising direction, existing data slicing based solutions suffer from tail latency. To overcome this limitation, this research introduces a novel communication-computation overlap technique to eliminate this tail latency in state of the art overlap methods for distributed LLM training. The aim of this technique is to effectively mitigate communication bottleneck of tensor parallelism and data parallelism for distributed training and inference. In particular, we propose a novel method termed \emph{CommFuse} that replaces conventional collective operations of reduce-scatter and all-gather with decomposed peer-to-peer (P2P) communication and schedules partitioned computations to enable fine-grained overlap. Our method provides an exact algorithm for reducing communication overhead that eliminates tail latency. Moreover, it presents a versatile solution compatible with data-parallel training and various tensor-level parallelism strategies, including TPSP and UP. Experimental evaluations demonstrate that our technique consistently achieves lower latency, superior Model FLOPS Utilization (MFU), and high throughput.
\end{abstract}


\section{Introduction}
\label{intro}
\begin{wrapfigure}{r}{0.5\textwidth}
    \centering
    \resizebox{\linewidth}{!}{%

    \begin{tikzpicture}[x=1cm,y=1cm,font=\sffamily]
          \definecolor{COMP.blue}{RGB}{0,170,230}
          \definecolor{rsred}{RGB}{255,0,0}
          \definecolor{overpink}{RGB}{244,170,170}
          \fill[COMP.blue] (0,0) rectangle (4.7,0.55);
          \fill[overpink] (4.7,0) rectangle (9.30,0.55);
          \fill[rsred]    (4.7,-0.35) rectangle (9.30,0);
          \node[font=\sffamily\tiny,align=center] at (2.3,0.275) {Compute};
          \node[font=\sffamily\tiny] at (6.425,0.275) {overhead};
          \node[font=\sffamily\tiny,align=center] at (7.15,-0.175) {RS};
          \node[font=\sffamily\small] at (4.05,-0.42) {No overlap};
         
          \begin{scope}[yshift=-1.45cm]
            \fill[COMP.blue] (0,0) rectangle (2.34,0.55);
            \fill[COMP.blue] (2.35,0) rectangle (4.7,0.55);
            \fill[overpink] (4.7,0) rectangle (7.05,0.55);
            \fill[rsred]    (2.35,-0.35) rectangle (4.69,0);
            \fill[rsred]    (4.70,-0.35) rectangle (7.05,0);
            \node[font=\sffamily\tiny,align=center] at (1.175,0.275) {Compute};
            \node[font=\sffamily\tiny,align=center] at (3.525,0.275) {Compute};
            \node[font=\sffamily\tiny] at (5.425,0.275) {overhead};
            \node[font=\sffamily\tiny,align=center] at (3.3,-0.175) {RS};
            \node[font=\sffamily\tiny,align=center] at (5.2,-0.175) {RS};
            \node[font=\sffamily\small] at (3.00,-0.72) {Data Slicing};
          \end{scope}
    
          \begin{scope}[yshift=-3.25cm]
            \foreach \i in {0,1,2,3} {
              \fill[COMP.blue] (1.2*\i,0) rectangle +(1.19,0.55);
              \node[font=\sffamily\tiny,align=center] at (1.2*\i+0.6,0.275) {Compute};
            }
            \fill[overpink] (4.8,0) rectangle (5.99,0.55);
            \node[font=\sffamily\tiny,align=center] at (5.42,0.275) {overhead};
        
            \foreach \i in {0,1,2,3} {
              \fill[rsred] (1.2+1.20*\i,-0.35) rectangle +(1.19,0.35);
              \node[font=\sffamily\tiny,align=center] at (1.2+1.20*\i+0.6,-0.175) {RS};
            }
            \node[font=\sffamily\small] at (3.0,-0.72) {Increasing the number of data slices};
          \end{scope}
        
          \begin{scope}[yshift=-5.35cm]
            \foreach \i in {0,1,2,3} {
              \fill[COMP.blue] (1.2*\i,0) rectangle +(1.19,0.55);
              \node[font=\sffamily\tiny,align=center] at (1.2*\i+0.6,0.275) {Compute};
            }
            \foreach \i in {0,1,2} {
              \fill[rsred] (1.2+1.20*\i,-0.35) rectangle +(1.19,0.35);
              \node[font=\sffamily\tiny,text=black] at (1.2+1.20*\i+0.55,-0.175) {P2P};
            }
            \node[font=\sffamily\small] at (3.0,-0.72) {CommFuse (\textbf{Ours}) };
          \end{scope}
        
    \end{tikzpicture}

    }
    \caption{A graphical example showing how the proposed solution can hide the communication overhead under computation and hence overcome the limitations of state of the art approaches.}
    \label{fig:overhead_comparison_abstract}
\end{wrapfigure}
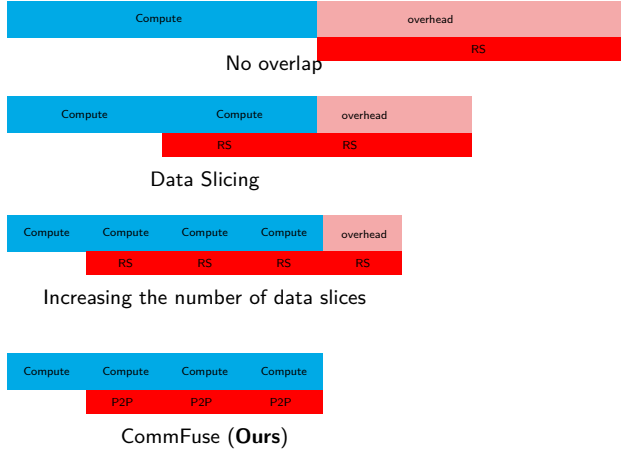

The rapid growth in the size of large language models has made distributed training and inference essential, requiring multi-dimensional parallelism across accelerators such as GPUs,TPUs and NPUs \cite{zeng2023distributed, li2024efficient, duan2024efficient}. These distributed strategies partition computational workloads across clusters of accelerators which require collective communication among the the accelerators, often resulting in substantial performance bottleneck from the communication overhead as revealed in both theoretical analysis and empirical studies evident from recent insights \cite{zeng2025distributed, amer2026distributed}. This communication overhead can result from both waiting for a communication to complete and synchronization requirement at the communication operators. As a direct impact, research in hiding communication over computation in distributed parallel strategies has become an active area of research for efficient AI. This research presents a communication-computation overlap method applicable to multiple parallelism strategies that can overcome the limitation of state of the art approaches with the benefit of potential to fully hide the communication overhead as pictorially shown in Figure~\ref{fig:overhead_comparison_abstract}.

Widely used parallelism strategies exhibit distinct communication patterns. Data Parallelism (DP)~\cite{goyal2017accurate,li2020pytorch} with weight sharding relies on reduce-scatter (RS) operations during the backward pass. Tensor Parallelism (TP)~\cite{Samuel59} incurs all-reduce (AR) operations, while Tensor Parallelism combined with sequence parallelism (TPSP)~\cite{korthikanti2023reducing} involves both RS and all-gather (AG) operations in the forward and backward passes. Ulysses Parallelism (UP)~\cite{jacobs2023deepspeed} uses all-to-all communication, for which no efficient computation-communication algorithm has been reported to date. These communication overheads can significantly limit training and inference efficiency when parallelization is employed.


Recent research addressed to overcome the communication overhead by hiding communication under computation. Previous works typically follows two broad categories of approaches: decomposing the data into smaller chunks vs. decomposing the collective communication algorithm. The former  direction works by decomposing the data into chunks and interleaving the computation of one chunk with an asynchronous communication of other chunk so that the computation and communication occurs in parallel. This type of approach is currently used in mainstream platforms, such as Megatron~\cite{shoeybi2019megatron,megatronlm2025github} and MindSpeed~\cite{mindspeedllm}. The other direction decomposes the collective communication to a set of different communication pattern, such as peer to peer, and decouple the interdependency to improve the overlap ratio. There has been application of these approach of TPUs~\cite{wang2022overlap}. In addition, fusing the kernels are used for efficient implementation for all type of approaches~\cite{chang2024flux,chen2024centauri,zheng2025tilelink}. 

Despite the recent progress, there still exist numerous limitation requiring further attention and room for improvement. The major drawbacks of decomposing data-based approaches is the tail overhead. Although the approaches split data and can overlap the communication of intermediate chunks when communication time is less than compute time, there is an overhead of communication time of last chunk. Again, making too small split to make to tail overhead smaller is not always beneficial for reasons of reduced arithmetic intensity and operations to become memory bound. Furthermore, inter-node collective communication is not efficient which limits the sweet spot of intra-layer model parallelism to be used across accelerators within a single node. Conversely, the major drawbacks of prior art following decomposing collective communication algorithm faces challenges in strict synchronization requirement at intermediate steps, and increased total communication volume. Although these types of approaches are expected to potentially achieve better overlap than the previous category, the above limitations hinder the use of the full potential. 

This work addresses several critical challenges in distributed model training and inference. First, it significantly reduces communication overhead in intra-layer model parallelism—including TP, TPSP, and UP—by overlapping computation and communication, making it applicable to a wide range of architectures such as Transformers, Mamba, and Hybrid models. This reduction directly improves training and inference latency, decreases time per output token, and increases overall system throughput, contributing to lower operational costs and greater energy efficiency. Moreover, by alleviating the synchronization and communication bottlenecks that typically constrain TP and TPSP to a single node, our approach enables intra-layer model parallelism to scale across multiple nodes. This extension supports the training and inference of extremely large models with higher parameters per layer, opening the door for TP/TPSP deployments beyond the single-node limit imposed by existing state-of-the-art methods.

To address the aforementioned challenges, we propose \emph{CommFuse}, a novel method that decomposes reduce-scatter (RS) and all-gather (AG) into peer-to-peer (P2P) communications and schedules partitioned computations for fine-grained overlap. We refer to these instantiations as \emph{FuseRS} and \emph{FuseAG}. Unlike prior approaches, \emph{CommFuse} provides an exact algorithm that eliminates tail overhead, supports data parallel (DP) training, and is compatible with tensor-level parallelism strategies such as TPSP and UP. Across multiple accelerators and model architectures, our method consistently improves Model FLOPS Utilization (MFU) resulting in reduced latency.

\emph{CommFuse} is a unified framework for fully overlapping communication and computation in distributed sequence models. First, we decompose the reduce-scatter collective into an asynchronous point-to-point (P2P) communication framework, eliminating intermediate synchronization and enabling flexible scheduling across participants. Second, we propose a rank-adaptive partial output scheduling strategy that prioritizes communication-dependent computations while deferring communication-free outputs, thereby mitigating tail latency. Third, we present the first solution that achieves full overlap between communication and computation, effectively eliminating communication overhead when communication latency is amortized by computation and minimizing it otherwise. Fourth, our design is unified across training and inference (including both prefill and decode phases), removing the need for separate parallelization strategies. Fifth, the proposed overlap mechanism is batch-size agnostic, maintaining efficiency even in small-batch or long-context regimes where prior batch-splitting approaches degrade. Finally, the method is orthogonal to attention implementations and remains fully compatible with optimized kernels such as Flash Attention.

The main contributions of this paper are:
\begin{itemize}
    \item \textbf{Eliminating tail overhead and synchronization constraints.} \emph{CommFuse} fully overlaps communication and computation for RS and AG, removing tail overhead and reducing intermediate synchronization requirements.  
    \item \textbf{Versatile parallelization.} The method efficiently supports DP, TP, UP across diverse models, enabling low-latency, high-MFU distributed training and inference.  
    \item \textbf{Robust and flexible.} \emph{CommFuse} is effective across varying batch sizes and context lengths and is compatible with techniques such as flash attention for transformer-based models.
\end{itemize}

\section{Related Works}
A central challenge in overlapping computation and communication lies in managing data dependencies. Weak dependencies, such as the all-reduce in the backward pass of data parallelism (DP), contrast with strong dependencies, such as reduce-scatter and all-reduce in tensor parallelism (TP/TPSP). Algorithms that integrate computation and communication must explicitly account for these dependency constraints, while optimized fused kernels can further benefit from low-level performance improvements. Existing research broadly follows two directions: (1) approximate or compressed communication (e.g., Flash Communication~\cite{li2024flash}), and (2) exact communication with improved decomposition and scheduling algorithms (e.g., Decompose~\cite{wang2022overlap}). Key directions in exact solutions focus on data decomposition and algorithm restructuring, often leveraging fused kernels for additional performance gains~\cite{wei2024communication}.

 Widely used frameworks, such as Megatron~\cite{megatronlm2025github} uses overlap of communication considering data slicing based approach. For DP, it allows to overlap the reduce-scatter of backward to overlap with the computation of forward of another data block. For TP/TPSP, it allows tiling of data into small chunks and overlap computation and communication of independent chunks. MindSpeed~\cite{mindspeedllm} also follows similar approach known as Ascend MC2. Yet another similar approach is Ascend CoC~\cite{ascendcoc}.

The direction of algorithm restructuring decompose the collective operation algorithm in addition to decomposing the data and schedules compute and communication for better overlapping. General  direction is to use graph transformation for transforming the fine grained computation to semantically equivalent graph with decoupled asynchronous instruction scheduling that allow higher overlapping. By breaking down blocking collective communication into a sequence of single-step non-blocking collectives,  a better overlap can be achieved. The Decompose~\cite{wang2022overlap} algorithm decomposes computation operations into finer-grained tasks and overlaps with the decomposed non blocking communications and executes then in a simple sequence to accumulate partial results.

Fine-grained implementation optimizations to improve compute–communication overlap often utilizes kernel fusion or scheduling. MiCS~\cite{zhang2022mics} reduces communication overhead via scale-aware partitioning, hierarchical communication, and 2-hop gradient synchronization. CoCoNet~\cite{jangda2022breaking} uses operator fusion with a compiler-based system to co-optimize computation and communication. Oases~\cite{li2023automated} introduces automated tensor model parallelism with fine-grained scheduling to maximize overlapping dependent operations. Flux~\cite{chang2024flux} over-decomposes operations into thread-block tasks and fuses them into larger kernels for fine-grained overlap. ISO~\cite{xiao2024iso} targets sequence-level overlap for LLM inference prefill stages. Concerto~\cite{cheng2025concerto} and TileLink~\cite{zheng2025tilelink} explore automated kernel optimization and scheduling to enhance overlap. These works provide complementary solutions, but are less directly related to algorithm-level decomposition.  
 
The proposed method of this research adopts algorithm-level decomposition and fusion to avoids strict synchronization constraints and tail overheads of prior methods, while presenting a versatile solution extending to multiple parallelization applications. This aligns closely with the algorithm restructuring approaches discussed above, while remaining orthogonal to the efficient kernel implementation techniques.

\section{Method}
The proposed \emph{CommFuse} builds upon the novel Fuse All Gather (\emph{FuseAG}) and Fuse Reduce Scatter (\emph{FuseRS}) primitives. These two algorithms can be directly instantiated for DP and TPSP, while a subtle extension enables support for UP. We begin by briefly reviewing the fundamentals of these parallelism strategies and their associated collective communication patterns, and then transition to a discussion of the proposed approach. Throughout this paper, we use the term \textit{rank} to denote a compute unit in a distributed setting; we also use the terms \textit{accelerator} or \textit{device} to refer to the physical instantiation of a compute rank.

\subsection{Preliminaries}
\paragraph{Parallelization.}
Zero Data Parallelism (Zero-DP) is a distributed training strategy in which model parameters, gradients, and optimizer states are partitioned across multiple accelerators to reduce memory redundancy while preserving data-parallel semantics. In this setting, the forward pass requires an all-gather operation to materialize the full set of model weights on each rank for computation, whereas the backward pass employs all-reduce or reduce-scatter operations to aggregate and distribute gradients across ranks efficiently. 

Tensor Parallelism (TP) distributes large model tensors across multiple accelerators by partitioning them and coordinating computation through collective communication primitives. Specifically, TP typically relies on all-reduce operations to combine partial results across ranks and may use all-gather to assemble intermediate activations when required, thereby enabling efficient computation and improved memory utilization. 

Tensor-Parallel Sequence Parallelism (TPSP) integrates Tensor Parallelism (TP) with Sequence Parallelism (SP), where SP partitions input sequences across ranks and commonly utilizes all-gather and reduce-scatter operations to manage activations and gradients across sequence partitions. Within this framework, SP is primarily applied to normalization layers, while TP is employed in the MLP and attention layers, collectively optimizing parallel execution and communication efficiency across model components. 

Ulysses Parallelism is a sequence-level parallelism strategy that partitions long input sequences across multiple GPUs and redistributes attention heads using all-to-all communication. This design enables efficient multi-GPU computation of attention over extremely long contexts by ensuring that each rank receives the necessary sequence segments and attention heads for localized computation while maintaining global consistency.

\begin{figure}
    \centering
    \includegraphics[width=0.7\linewidth]{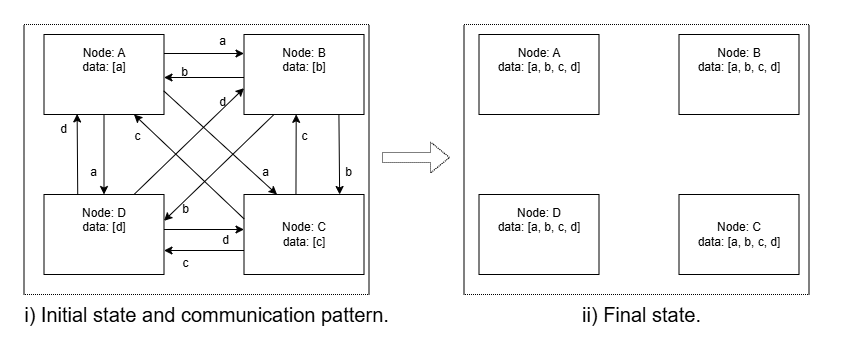}
    \caption{All Gather (AG) operation showing initial state and communication pattern on the left and final state on the right.}
    \label{fig:ag}
\end{figure}

\begin{figure}
    \centering
    \includegraphics[width=0.95\linewidth]{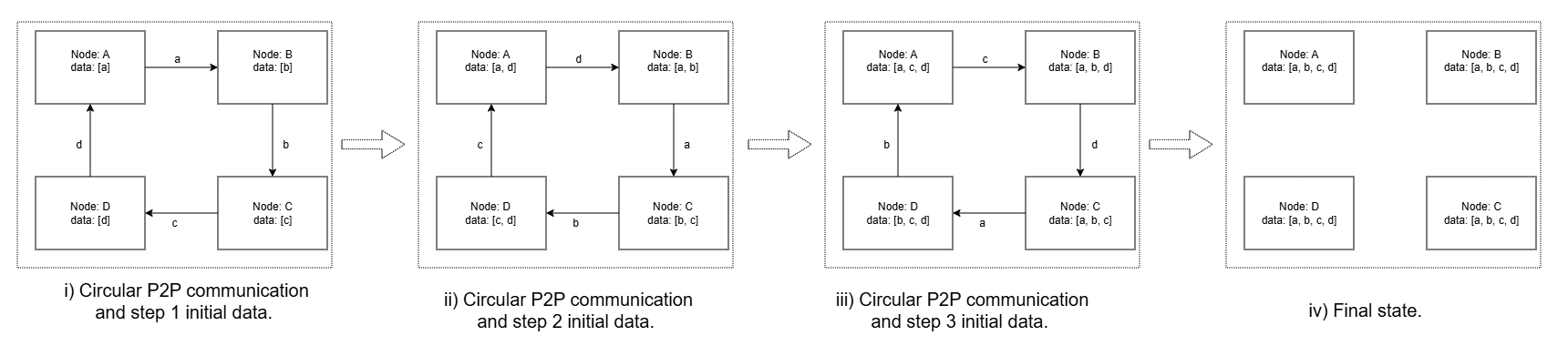}
    \caption{Ring implementation of All-Gather operation showing example for four compute ranks.}
    \label{fig:ag-ring}
\end{figure}

\paragraph{Common collective communications.}
The All-Gather operation collects data from all participating devices such that, upon completion, every device holds an identical copy of the aggregated data. For example, consider a system with four ranks, where each rank initially contains a distinct data element: rank 0 holds $a$, rank 1 holds $b$, rank 2 holds $c$, and rank 3 holds $d$. After the All-Gather operation, each rank possesses the complete set of elements, i.e., $(a, b, c, d)$. A visual illustration is provided in Figure~\ref{fig:ag}. In practice, the All-Gather (AG) collective communication is commonly implemented using a multi-step ring-based communication pattern, as illustrated in Figure~\ref{fig:ag-ring}.

\begin{figure}
    \centering
    \includegraphics[width=0.7\linewidth]{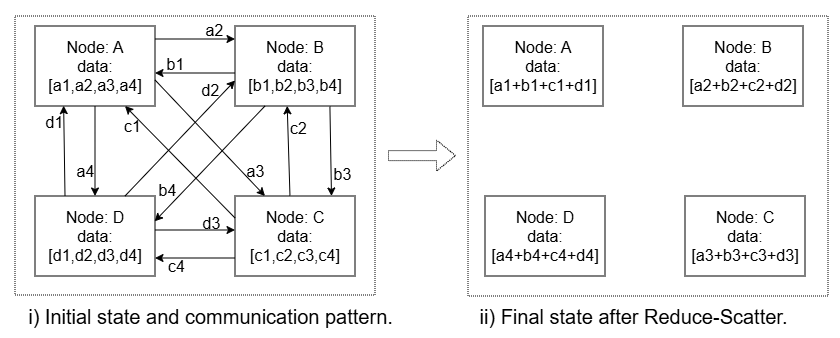}
    \caption{Reduce-Scatter (RS) operation showing initial state and communication pattern on the left and final state on the right.}
    \label{fig:rs}
\end{figure}

The Reduce-Scatter (RS) operation, in contrast, partitions data across ranks by slice index and redistributes these slices such that each rank aggregates the corresponding slice from all other ranks. Consequently, after the RS operation, each compute rank retains a unique portion of the globally reduced data, consisting of the aggregation of that specific slice across all ranks. For example, consider four compute ranks initially holding the data sequences $(a_0, a_1, a_2, a_3)$, $(b_0, b_1, b_2, b_3)$, $(c_0, c_1, c_2, c_3)$, and $(d_0, d_1, d_2, d_3)$, respectively. When applying RS with summation as the reduction operator, each rank receives the aggregated slice corresponding to its index, resulting in $(a_0 + b_0 + c_0 + d_0)$, $(a_1 + b_1 + c_1 + d_1)$, $(a_2 + b_2 + c_2 + d_2)$, and $(a_3 + b_3 + c_3 + d_3)$ across the four ranks. A visual illustration of this process is shown in Figure~\ref{fig:rs}.

\begin{figure}
    \centering
    \includegraphics[width=0.95\linewidth]{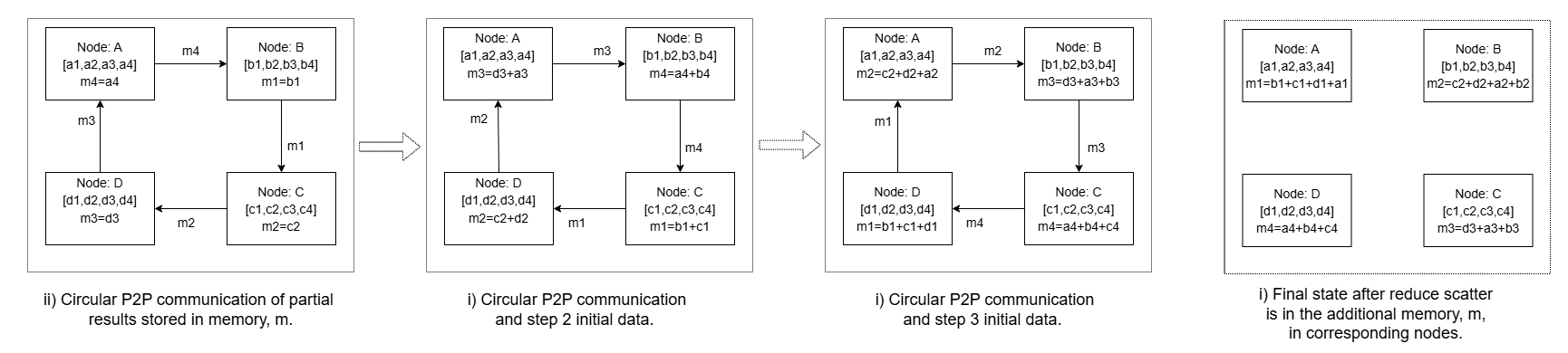}
    \caption{Ring implementation of Reduce-Scatter operation showing example for four compute ranks.}
    \label{fig:rs-ring}
\end{figure}

Similarly, the Reduce-Scatter (RS) collective communication is typically realized through a multi-step ring communication scheme, as shown in Figure~\ref{fig:rs-ring}. The proposed approach leverages this ring-based implementation of collective communication, although alternative bidirectional peer-to-peer communication patterns may also be employed.

\subsection{\emph{CommFuse}}
The proposed method overlaps collective communication with computation to hide the overhead of all-gather and reduce-scatter operations in TPSP, as well as all-to-all communication in UP. The approach decomposes computations into multiple iterative steps and collective communications into peer-to-peer exchanges. A rank-wise adaptive scheduling is employed to maximize the overlap between computation and communication, ensuring that tail computations—such as the sequence slice remaining on the current rank—do not require further communication. This strategy effectively conceals communication latency under computation. Consequently, a sequence of length $S$ is partitioned into $m \times N$ slices for computation, where $m \in \{1, 2, \dots\}$ defines the slicing granularity and $N$ denotes the number of compute nodes. For simplicity, a typical value of $m=1$ is used throughout this work, and $m$ is omitted from the algorithm descriptions unless stated otherwise.

The overall algorithm for communication overlaps for tensor level model parallelism includes ‘Fuse All Gather’ and ‘Fuse Reduce Scatter’ which uses overlapped computation and replaces conventional collective communication of ’All Gather’ and ‘Reduce Scatter’. Both of the algorithm uses the same principle of breaking down the collective communication into an efficient decomposed algorithm of a sequence of peer-to-peer communications, and schedule computation of partial outputs in a way that eliminate tail overhead. It is achieved by considering the nature of communication and existing data in each compute node where the algorithms for all-gather and reduce-scatter differs. All gather communication begins with a slice of data in each compute nodes and gathers data from multiple ranks for compute. Hence, the all gather commutes data before the computation. Considering this case of all-gather, we schedule the compute of existing data slice to happen first and then we schedule compute of received. On the other hand, reduce-scatter communication happens of computed outputs, and hence, we schedule the compute of the data slice that will remain in the current compute node to happen as the last step. In the following, we present these as two different algorithm both of which built on the same principle with particular order of compute and communication depending on the collective communication nature. We describe embodiment of these algorithms to particular layer types (such as MLP, Attention and Mamba) in following Section.

\begin{figure}
    \centering
    \includegraphics[width=0.95\linewidth]{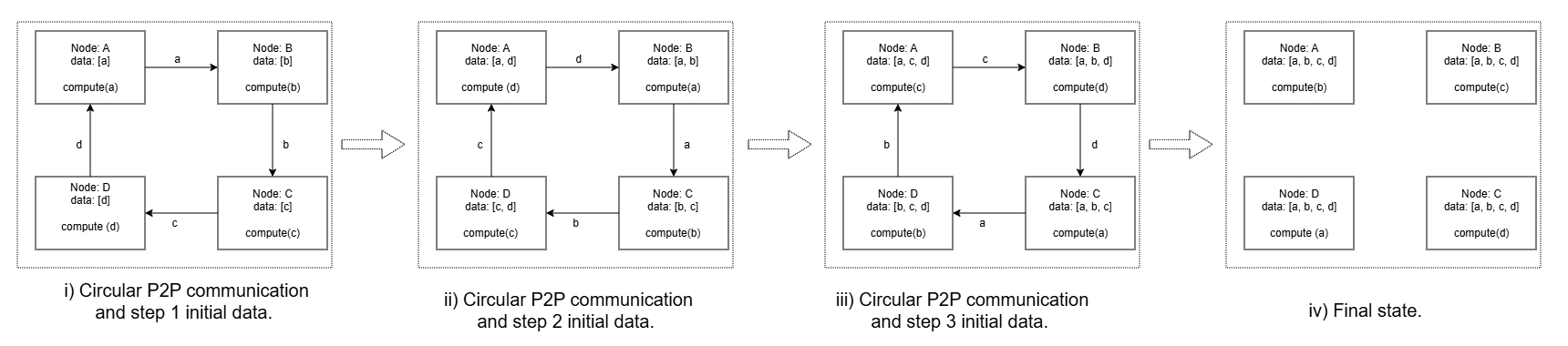}
    \caption{Four-rank example of \emph{Fuse All Gather}, showing the ring-ordered breakdown of the collective into peer-to-peer communication steps with interleaved partial-output computation to achieve compute–communication overlap.}
    \label{fig:agwithringoverlap}
\end{figure}

\paragraph{Fuse All Gather.}
An algorithm for ``Fuse All Gather'' is presented in Algorithm~\ref{algo:fuseag}. This instantiation considers a ring-based communication in the clockwise direction; however, alternative instantiations are possible by changing the communication direction while preserving all key principles. These variations are considered implementation-specific extensions of the main approach. The symbols used follow conventional meanings: $B$ denotes batch size, $S$ sequence length, $D$ model dimension, $X$ input, $O$ output, $r$ tensor parallel rank, $T$ tensor parallel world size, and $N$ the number of computation decompositions. When $m=1$, $N=T$. The function \texttt{compute\_partial\_output} represents a high-level abstraction of the computation of a partial output corresponding to a decomposed slice or chunk of the input; for example, $O_j^r$ denotes the partial output at rank $r$ of sequence chunk $j$. In specific implementations, \texttt{compute\_partial\_output} is replaced with the corresponding operations, typically a column-shard linear projection. When the peer-to-peer transfer time $T_{\text{comm}}$ of an output chunk $O_j^r$ of shape $B \times S/N \times D$ is less than its compute time $T_{\text{comp}}$, the communication overhead can be fully hidden; otherwise, this algorithm achieves the maximum possible reduction of overhead. Further optimization is possible at the implementation level by developing fused kernels that integrate computation and communication. Figure~\ref{fig:agwithringoverlap} illustrates a four-node configuration executing \emph{Fuse Reduce Scatter}, where the Reduce Scatter collective is decomposed into a sequence of peer-to-peer communication steps with interleaved computation arranged to maximize compute–communication overlap. Although alternative schedules are possible, including bidirectional peer-to-peer decompositions with different ordering strategies, we present the ring-based ordering to clearly demonstrate the core mechanism.


\begin{algorithm}
\caption{FuseAllGather}
\label{algo:fuseag}
\begin{algorithmic}[1]
\State \textbf{Input:} Rank $r$ Sequence $X^r \in \mathbb{R}^{B \times \frac{S}{T} \times D}$
\State \textbf{Parameters:} Rank $r$ Column Shard Parameters, $W^r$
\State \textbf{Output:} Output Tensor $O^r$
\State $r \gets current\_rank $ \Comment{Rank of current computing node}
\State $T \gets group\_size $ \Comment{Number of nodes in the group}
\State $N \gets T$ \Comment{ Number of sequence slices}
\State $O^r_{s \mid s \in \{0, \dots, N-1\}} \gets$ \texttt{init\_output\_sequence\_list} \Comment{ Placeholder for output}
\State $B^r_{in} \in \mathbb{R}^{B \times \frac{S}{T} \times D} \gets$ \texttt{init\_receive\_buffer} \Comment{Buffer to receive data}
\State $B^r_{out} \in \mathbb{R}^{B \times \frac{S}{T} \times D} \gets$ \texttt{init\_send\_buffer} \Comment{Buffer for sending data}
\State $B^r_{out} \gets X^r$ \Comment{Begin with the existing sequence slice}
\For{$i = 0$ to $N-1$}  \Comment{ N iterations}
    \State $j = (r+1)\mod N $ \Comment{ Send data to next rank}
    \State $k = (r-1+N)\mod N$ \Comment{Receive data from previous rank}
    \State $l = (r-i+N)\mod N $ \Comment{Compute slice index} 
    \If{ $i < N-1 $}  \Comment{ Decomposed communication} 
        \State $p2p\_send(data=B^r_{out},rank=j,async )$ 
        \State $p2p\_recv(data=B^r_{in},rank=k,async )$ 
    \EndIf
    \State $O^r_s[l] \gets compute\_partial\_output(B^r_{out},W^r)$ \Comment{ Decomposed compute}
    \State \texttt{Wait for communication to complete}  
    \State $B^r_{out} \gets B^r_{in}$ \Comment{Forward the received data slice to next rank}
 \EndFor
\State $O^r \gets \texttt{Concatenate\_Sequences} (O^r_s) $ \Comment{Concatenate the sequence slices}
\State \Return $O^r$ 
\end{algorithmic}
\end{algorithm}

\paragraph{Fuse Reduce Scatter.}

\begin{figure}
    \centering
    \includegraphics[width=0.95\linewidth]{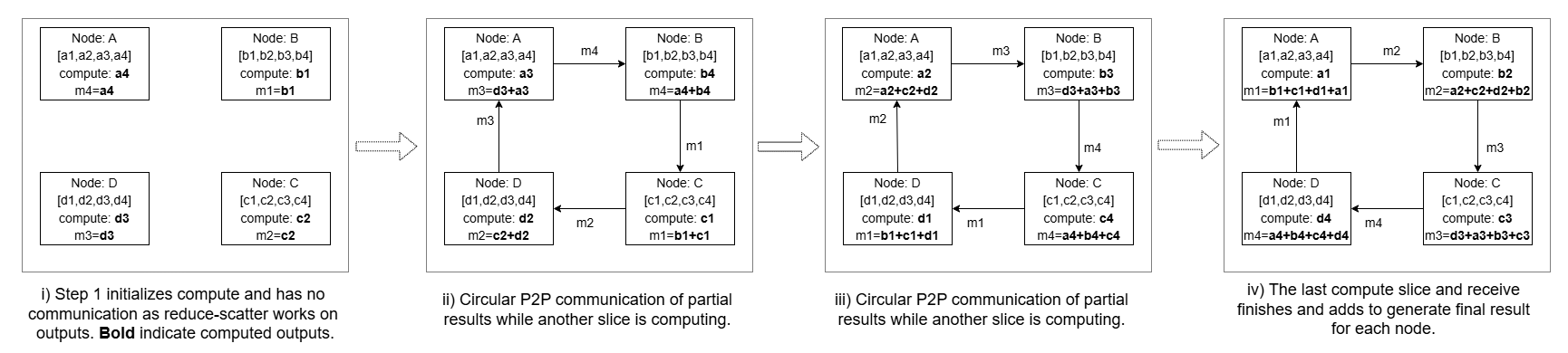}
    \caption{Four-rank example of \emph{Fuse Reduce Scatter}, showing the ring-ordered breakdown of the Reduce Scatter collective into peer-to-peer communication steps with interleaved partial-output computation to achieve compute–communication overlap.}
    \label{fig:rswithringoverlap}
\end{figure}

The \emph{Fuse Reduce Scatter} method is shown in Algorithm~\ref{algo:fuse_rs}. Similar to the previous algorithm, the symbols follow their conventional meanings: $B$ denotes the batch size, $S$ the sequence length, $D$ the model dimension, $X$ the input, $O$ the output, $r$ the tensor-parallel rank, $T$ the tensor-parallel world size, and $N$ the number of decompositions of the computation; when $m = 1$, we have $N = T$. The function \texttt{compute\_partial\_output} abstracts the decomposed computation that produces a partial output for a specific slice of the input, where $O_j^r$ denotes the partial output at rank $r$ for sequence chunk $j$. In most cases this corresponds to a row-sharded linear projection, though additional operations may be incorporated to enhance compute–communication overlap depending on the layer type (e.g., MLP, Attention, Mamba), and in concrete implementations it is replaced with the appropriate layer-specific computation. When the point-to-point transfer time $T_{\text{comm}}$ of an output chunk $O_j^r$ of shape $B \times (S/N) \times D$ is lower than the compute time $T_{\text{comp}}$ for that chunk, communication can be fully hidden; when $T_{\text{comp}} < T_{\text{comm}}$, the algorithm still achieves the maximum possible reduction in overhead. Depending on hardware capabilities, the aggregation over the $N$ buffers may be decomposed and performed incrementally within the loop, and further optimizations—such as fused compute–communication kernels—can be applied at the implementation level. Figure~\ref{fig:rswithringoverlap} illustrates a four-node configuration executing \emph{Fuse Reduce Scatter}, where the Reduce-Scatter collective is decomposed into a sequence of peer-to-peer communication steps with interleaved computation arranged to maximize compute–communication overlap. Although alternative schedules are possible, including bidirectional peer-to-peer decompositions with different ordering strategies, we present the ring-based ordering to clearly demonstrate the core mechanism.

\begin{algorithm}
\caption{FuseReduceScatter}
\label{algo:fuse_rs}
\begin{algorithmic}[1]
\State \textbf{Input:} Rank $r$ Sequence  $X^r \in \mathbb{R}^{B \times S \times \frac{D}{T}}$
\State \textbf{Parameters:} Rank $r$ Row Shard Projection Parameters, $W^r$
\State \textbf{Output:} Output Tensor $O^r$
\State $r \gets current\_rank $ \Comment{Rank of current computing node}
\State $T \gets group\_size $ \Comment{Number of nodes in the group}
\State $N \gets T$ \Comment{ Number of sequence slices}
\State $O^r_{s \mid s \in \{0, \dots, N-1\}} \gets$ \texttt{init\_sequence\_as\_sequence\_list}$(X^r, N)$

\State $B^r_{in} \in \mathbb{R}^{B \times \frac{S}{T} \times D} \gets$ \texttt{init\_receive\_buffer} \Comment{Buffer to receive data}
\State $B^r_{out} \in \mathbb{R}^{B \times \frac{S}{T} \times D} \gets$ \texttt{init\_send\_buffer} \Comment{Buffer for sending data}

\For{$i = 0$ to $N-1$}  \Comment{ N iterations}
    \State $j = (r+1)\mod N $ \Comment{ Send data to next rank}
    \State $k = (r-1+N)\mod N$ \Comment{Receive data from previous rank}
    \State $l = (r-i-1+N)\mod N $ \Comment{Compute slice index} 

    \State $O^r \gets compute\_partial\_output(X_s[l],W^r)$ \Comment{ Decomposed compute}
    \If{ $i > 0 $}  \Comment{ Aggregate result from second to last steps}
        \State $O^r \gets B^r_{in} + O^r$  
    \EndIf    
    \If{ $i < N-1 $}  \Comment{ Decomposed communication}
        \State $B^r_{out} \gets O^r$
        \State $p2p\_send(data=B^r_{out},rank=j,async )$ 
        \State $p2p\_recv(data=B^r_{in},rank=k,async )$ 
    \EndIf
 \EndFor
\State \Return $O^r$ 
\end{algorithmic}
\end{algorithm}

\section{Architectural Instantiations}

The presented solution can be applied for a wide range of layer types including MLP, Attention~\cite{vaswani2017attention}, and Mamba~\cite{gu2024mamba,waleffe2024empirical,lahoti2026mamba} layers. Using the ‘Fuse All Gather’ and ‘Fuse Reduce Scatter’, we built ‘Fuse TPSP’ for TPSP with full communication/compute overlap. We also apply the key principles of the method into ‘Fuse All-to-All’ for Ulysses Parallel Attention (UP). In the following we discuss the detailed instantiation for each of these layer types.

\begin{algorithm}
\caption{FuseReduceScatter:RowParallelProjection}
\label{algo:fuserowparallel}
\begin{algorithmic}[1]
\State \textbf{Input:} Rank $r$ Sequence  $X^r \in \mathbb{R}^{B \times S \times \frac{D}{T}}$
\State \textbf{Parameters:} Rank $r$ Row Shard Projection Parameters, $W^r_o \in \mathbb{R}^{\frac{D}{T} \times D}$
\State \textbf{Output:} Output Tensor $O^r$
\State $r \gets current\_rank $ \Comment{Rank of current computing node}
\State $T \gets group\_size $ \Comment{Number of nodes in the group}
\State $N \gets T$ \Comment{ Number of sequence slices}
\State $O^r_{s \mid s \in \{0, \dots, N-1\}} \gets$ \texttt{init\_sequence\_as\_sequence\_list}$(X^r, N)$

\State $B^r_{in} \in \mathbb{R}^{B \times \frac{S}{T} \times D} \gets$ \texttt{init\_receive\_buffer} \Comment{Buffer to receive data}
\State $B^r_{out} \in \mathbb{R}^{B \times \frac{S}{T} \times D} \gets$ \texttt{init\_send\_buffer} \Comment{Buffer for sending data}

\For{$i = 0$ to $N-1$}  \Comment{ N iterations}
    \State $j = (r+1)\mod N $ \Comment{ Send data to next rank}
    \State $k = (r-1+N)\mod N$ \Comment{Receive data from previous rank}
    \State $l = (r-i-1+N)\mod N $ \Comment{Compute slice index} 
    \State $O^r_j \gets W^r_o * X_s[j]$ \Comment{ Modified decomposed compute}
    \If{ $i > 0 $}  \Comment{ Aggregate result from second to last steps}
        \State $O^r \gets B^r_{in} + O^r$  
    \EndIf    
    \If{ $i < N-1 $}  \Comment{ Decomposed communication}
        \State $B^r_{out} \gets O^r$
        \State $p2p\_send(data=B^r_{out},rank=j,async )$ 
        \State $p2p\_recv(data=B^r_{in},rank=k,async )$ 
    \EndIf
 \EndFor
\State \Return $O^r$ 
\end{algorithmic}
\end{algorithm}

\subsection{Row Parallel Projection}
The implementation of the ``Fuse Reduce Scatter'' operation in Algorithm~\ref{algo:fuse_rs} is realized via a row-shard linear projection. We further consider alternative implementations of ``Fuse Reduce Scatter'' that enable the overlap of computation and communication, such as in the context of attention computation. Additionally, we examine an Ulysses Parallelism-specific all-to-all variant, termed ``Fuse All-to-All,'' as another embodiment for the attention mechanism. Given that these constitute non-trivial extensions of Algorithm~\ref{algo:fuse_rs}, we provide a more detailed discussion of these implementations. We first present the generic implementation of ``Fuse Reduce Scatter'' using row-shard linear projection while we also present some additional layer specific implementation in later section. Hence, in this generic row parallel projection implementation, the $compute\_partial\_output$ is implemented as a row sharded projection layer similart to megatron style row parallel projection with the equation: $O_j^r \gets W_o^r \times X_s[j]$. In this equation, $O_j^r$ is partial output from rank $i$ for index $j$, $W_o^r$ is row sharded linear projection weight parameters, and $X_s[j]$ is a slice of input partitioned in sequence dimension. Since generic, this row parallel projection is applicable to all type of layer which has a tensor parallel sharded projection layer followed by a reduce-scatter, such as MLP, Attention and Mamba. The detail algorithm of this embodiment is shown in Algorithm~\ref{algo:fuserowparallel}.

\begin{algorithm}
\caption{FuseReduceScatter:QuerySplitAttention}
\label{algo:querysplitattention}
\begin{algorithmic}[1]
\State \textbf{Input:} $Q^r,K^r,V^r \in \mathbb{R}^{B \times \frac{a}{T} \times S \times D_h }$ \Comment{$\texttt{Head dim}, D_h = \frac{D}{a}$}
\State \textbf{Parameters:} Rank $r$ Row Shard Projection Parameters, $W^r_o \in \mathbb{R}^{\frac{D}{T} \times D}$
\State \textbf{Output:} Output Tensor $O^r$

\State $r \gets current\_rank $ \Comment{Rank of current computing node}
\State $T \gets group\_size $ \Comment{Number of nodes in the group}
\State $N \gets T$ \Comment{ Number of sequence slices}
\State $Q^r_{s \mid s \in \{0, \dots, N-1\}} \gets$ \texttt{init\_sequence\_as\_sequence\_list}$(Q^r, N)$

\State $B^r_{in} \in \mathbb{R}^{B \times \frac{S}{T} \times D} \gets$ \texttt{init\_receive\_buffer} \Comment{Buffer to receive data}
\State $B^r_{out} \in \mathbb{R}^{B \times \frac{S}{T} \times D} \gets$ \texttt{init\_send\_buffer} \Comment{Buffer for sending data}

\For{$i = 0$ to $N-1$}  \Comment{ N iterations}
    \State $j = (r+1)\mod N $ \Comment{ Send data to next rank}
    \State $k = (r-1+N)\mod N$ \Comment{Receive data from previous rank}
    \State $l = (r-i-1+N)\mod N $ \Comment{Compute slice index} 

    \State $attention\_scores^r \gets Q^r_s[l] * {K^r}^T$ \Comment{Attention Score for Query Slice}
    \State $attention\_weight^r \gets Softmax(attention\_scores^r)$ \Comment{ For a Query Slice}
    \State $\tilde{O^r} \gets attention\_weight^r * V^r$ \Comment{ Attention Output for Query Slice}
    \State $O^r \gets W^r_o * \tilde{O^r}$ \Comment{Decomposed Projection}

    \If{ $i > 0 $}  \Comment{ Aggregate result from second to last steps}
        \State $O^r \gets B^r_{in} + O^r$  
    \EndIf
    
    \If{ $i < N-1 $}  \Comment{ Decomposed communication}
        \State $B^r_{out} \gets O^r$
        \State $p2p\_send(data=B^r_{out},rank=j,async )$ 
        \State $p2p\_recv(data=B^r_{in},rank=k,async )$ 
    \EndIf
 \EndFor
\State \Return $O^r$ 
\end{algorithmic}
\end{algorithm}

\subsection{Column Parallel Projection}
The implementation of the ``Fuse All Gather'' operation in Algorithm~\ref{algo:fuseag} is realized via a column-shard linear projection. This column-shard linear projection is employed as the initial layer of the MLP, the QKV projection within the Attention mechanism, and the in-projection of the Mamba layer. In practice this can be used as an efficient alternative to the column parallel projection in Megatron-LM~\cite{megatronlm2025github}. Analogous implementations can be applied to any other layer type that leverages a column-shard linear projection. The proposed implementation is achieved by instantiating the \texttt{compute\_partial\_output} routine utilizing the column-shard linear projection. Since the procedure is intuitive, we omit an explicit algorithmic description for this case.

\subsection{Query Split Attention}
An attention specific implementation in addition to the previously mentioned ‘row parallel embodiment’ in the output projection is decomposing attention computation with splitting the queries into slices. This implementation splits the query, Q, of attention into N slices/chunks where N can be equal or a multiple of group size (assuming m=1 for simplicity of explanation). And then it iteratively computes outputs for each query chunk, $Q_i$, for different attention heads at different TP ranks. The outputs corresponding to each Query slice is communicated after each iteration using peer to peer communication. Hence, the $compute\_partial\_output$ is implemented as computing attention of a slice of query with the equations: $scores^r \gets Q_s^r [j] \times (K^r)^T$;  $weight_j^r \gets Softmax(scores_j^r)$; $O_j^r \gets weight_j^r \times V^r $; $O_j^r \gets W_o^r \times  (O_j^r )$. The slices scheduling is done as mentioned in the generic algorithm. This results in more computation to be overlapped with the communication where only linear projection computation is not sufficient for full overlap, making suitable for high flops low bandwidth hardware. Yet this implementation still remains orthogonal to flash-attention. Details are shown in in Algorithm~\ref{algo:querysplitattention}.

\subsection{Attention UP}
All-to-all communication in Ulysses Parallelism is a collective operation that redistributes activation tensors across accelerators. At the onset of attention computation, an all-to-all operation applied to $Q$, $K$, and $V$ assigns each accelerator a subset of attention heads spanning the full input sequence, while computation is restricted to non-overlapping subsets of attention heads, thereby enabling sequence parallelism. Following attention computation, a second all-to-all operation restores the original sequence partitioning by redistributing to each accelerator a subset of the sequence, now enriched by contributions from all attention heads. The final all-to-all operation can further leverage the proposed approach, termed \emph{Fuse All-to-All} (Algorithm~5), wherein the \texttt{compute\_partial\_output} routine is defined as the computation of attention over a slice of the query without the output projection, i.e., $\text{scores}_{j}^{r} \leftarrow Q_{s}^{r}[j](K^{r})^{\top}$, $\text{weight}_{j}^{r} \leftarrow \text{Softmax}(\text{scores}_{j}^{r})$, and $\tilde{O}_{j}^{r} \leftarrow \text{weight}_{j}^{r}V^{r}$. Notably, the final output projection is not sharded in Ulysses Parallelism and is therefore excluded from the overlapped computation; further details are provided in Algorithm~\ref{algo:fuseup}.

\begin{algorithm}
\caption{FuseAllToAll:UlyssesParallelAttention}
\label{algo:fuseup}
\begin{algorithmic}[1]
\State \textbf{Input:} $Q^r,K^r,V^r \in \mathbb{R}^{B \times \frac{a}{T} \times S \times D_h}$ \Comment{$\texttt{Head dim}, D_h = \frac{D}{a}$}
\State \textbf{Parameters:} None, the overlap happens with attention compute
\State \textbf{Output:} Output Tensor $O^r$
\State $r \gets current\_rank $ \Comment{Rank of current computing node}
\State $T \gets group\_size $ \Comment{Number of nodes in the group}
\State $N \gets T$ \Comment{ Number of sequence slices}

\State $Q^r_{s \mid s \in \{0, \dots, N-1\}} \gets$ \texttt{init\_sequence\_as\_sequence\_list}$(Q^r, N)$

\State $B^r_{out} \in \mathbb{R}^{B \times \frac{S}{T} \times D} \gets$ \texttt{init\_send\_buffer} \Comment{Buffer for sending data}
\State $O^r_{s \mid s \in \{0, \dots, N-1\}} \gets$ \texttt{init\_output\_list} \Comment{Placeholder for output and receive data}
\For{$i = 0$ to $N-1$}  \Comment{ N iterations}
    \State $ j \gets (r+i+1)\%N$ \Comment{Send rank}
    \State $ k \gets (r-i-1+N)\%N$ \Comment{Receive rank}
    \State $l \gets j$ \Comment{ Compute slice index}
    \State $attention\_scores^r_l \gets Q^r_s[l] * \texttt{transpose} (K^r)$ \Comment{For current query slice}
    \State $attention\_weight^r_l \gets Softmax(attention\_scores^r_l)$ 
    \State $O^r_l \gets attention\_weight^r_l * V^r$ 

    \If{ $i < N-1 $}
        \State $B^r_{out} \gets O^r_l $
        \State $p2p\_send(data=B^r_{out},rank=j,async )$ 
        \State $p2p\_recv(data=O^r[k],rank=k,async )$ 
    \Else 
        \State $O^r[l] \ \gets O^r_l$ 
    \EndIf
 \EndFor
\State \texttt{Wait for all communication to complete} 
\State $O^r \gets$ \texttt{Concatenate\_Feature\_Dimension}$(O^r_s)$  
\end{algorithmic}
\end{algorithm}

\section{Evaluation}
To empirically verify the effectiveness of the \emph{CommFuse} approach, we adopt \emph{FuseRS} as a representative collective communication primitive. We investigate its application in conjunction with tensor parallelism and sequence parallelism (TPSP). Specifically, we first conduct an empirical validation of the \emph{FuseRS} implementation, focusing on row-parallel linear layers within the MLP and query-split attention within the attention layer.


\paragraph{Sanity Check.}
We first perform a sanity check to validate the application of TPSP with the proposed row-parallel linear layers and query-split attention. To this end, we compare the outputs of layers configured with TPSP against those of the same layers without TPSP. Both configurations produce consistent outputs, thereby passing the sanity check.

\paragraph{MLP Layer.}
We first implement the MLP layer using the proposed row-parallel linear projection with \emph{Fuse Reduce Scatter} (FuseRS) as a core building block. For the column-parallel linear projection, we retain the conventional design, similar to that used in Megatron, as our objective in this experiment is to isolate and evaluate the benefit of a single component. We compare our method against a baseline implementation without overlap, following a data-decomposition approach, across varying sequence lengths. 

\begin{table}[t]
\centering
\caption{Performance comparison of different approaches under configuration: $d_{\text{model}}=4096$, tensor parallelism (TP)$=4$, input tensor shape $(b,s,d)=(32,4096,4096)$, and averaged over 10 iterations for forward pass to the layer.}
\label{tab:latency_mlp}
\resizebox{0.99\linewidth}{!}{
\begin{tabular}{l |p{2.3cm} p{2.3cm} p{2.3cm} p{2.2cm} p{2.2cm}}
\toprule
\textbf{Algorithm} 
& \centering\textbf{Forward Compute Time (ms)} 
& \centering\textbf{Communication Overhead (ms)} 
& \centering\textbf{End-to-End Latency (ms)} 
& \centering\textbf{Overhead Reduction (\%)} 
& \centering\textbf{Latency Reduction (\%)} \tabularnewline
\midrule
Vanilla Baseline        & \centering 78.5 & \centering 43.8 & \centering 122.3 & \centering --   & \centering --   \tabularnewline
Data Slicing Approach   & \centering 76.9 & \centering 7.1  & \centering 83.9  & \centering 83.9 & \centering 31.4 \tabularnewline
Ours                    & \centering 77.0 & \centering 0.1  & \centering 77.1  & \centering 99.8 & \centering 36.9 \tabularnewline
\bottomrule
\end{tabular}
}
\end{table}

Table~\ref{tab:latency_mlp} presents the performance comparison across different approaches. The vanilla baseline incurs a substantial communication overhead of 43.8~ms per iteration, leading to an end-to-end latency of 122.3~ms. The data slicing approach significantly reduces the communication overhead to 7.1~ms, achieving an 83.9\% reduction and lowering the overall latency to 83.9~ms (31.4\% reduction). In contrast, our proposed method further reduces the communication overhead to 0.1~ms, corresponding to a 99.8\% reduction, and achieves the lowest end-to-end latency of 77.1~ms, yielding a 36.9\% improvement over the baseline. 

Compared to the data slicing approach, our method provides an additional latency reduction of 8.1\%, computed as $(83.9 - 77.1)/83.9 \times 100\%$. Overall, our method is able to hide over 99\% of the communication overhead, compared to approximately 84\% for the data slicing approach. From a theoretical perspective, the proposed approach is expected to fully hide (i.e., 100\%) the communication overhead; we expect which could potentially achievable using a fully fused kernel implementation. These results highlight the effectiveness of the proposed method in overlapping communication with computation under realistic experimental settings.

\begin{figure}[t]
\centering
\begin{tikzpicture}
\begin{axis}[
    name=commplot,
    width=0.45\linewidth,
    height=0.45\linewidth,
    xlabel={Sequence Length (xK)},
    ylabel={Comm. Overhead Reduction (\%)},
    xmode=log,
    log ticks with fixed point,
    xtick={1,2,4,8,16},
    ymin=0, ymax=105,
    ymajorgrids=true,
    grid style=dashed,
    legend style={at={(0.11,1.33)}, anchor=north west, font=\small},
]

\addplot[blue, mark=o, thick] coordinates {
    (1,77.9)
    (2,78.9)
    (4,83.9)
    (8,82.7)
    (16,85.2)
};
\addlegendentry{Data Slicing}

\addplot[cyan, mark=square, thick] coordinates {
    (1,99.7)
    (2,99.7)
    (4,99.8)
    (8,99.8)
    (16,99.8)
};
\addlegendentry{Ours}

\end{axis}

\begin{axis}[
    at={(commplot.west)}, 
    xshift=6.5cm,         
    anchor=west,
    width=0.45\linewidth,
    height=0.45\linewidth,  
    xlabel={Sequence Length (xK)},
    ylabel={Latency Improvement (\%)},
    xmode=log,
    log ticks with fixed point,
    xtick={1,2,4,8,16},
    ymin=0, ymax=105,
    ymajorgrids=true,
    grid style=dashed,
    legend style={at={(0.11,1.33)}, anchor=north west, font=\small},
]

\addplot[blue, mark=o, thick] coordinates {
    (1,29.9)
    (2,35.0)
    (4,31.4)
    (8,29.5)
    (16,33.5)
};
\addlegendentry{Data Slicing}

\addplot[cyan, mark=square, thick] coordinates {
    (1,35.8)
    (2,40.2)
    (4,36.9)
    (8,35.0)
    (16,38.8)
};
\addlegendentry{Ours}

\end{axis}
\end{tikzpicture}
\caption{Comparison across sequence lengths. Left: Communication Overhead Reduction. Right: End-to-End Latency Improvement.}
\label{fig:comm_reduction_scaling}
\end{figure}
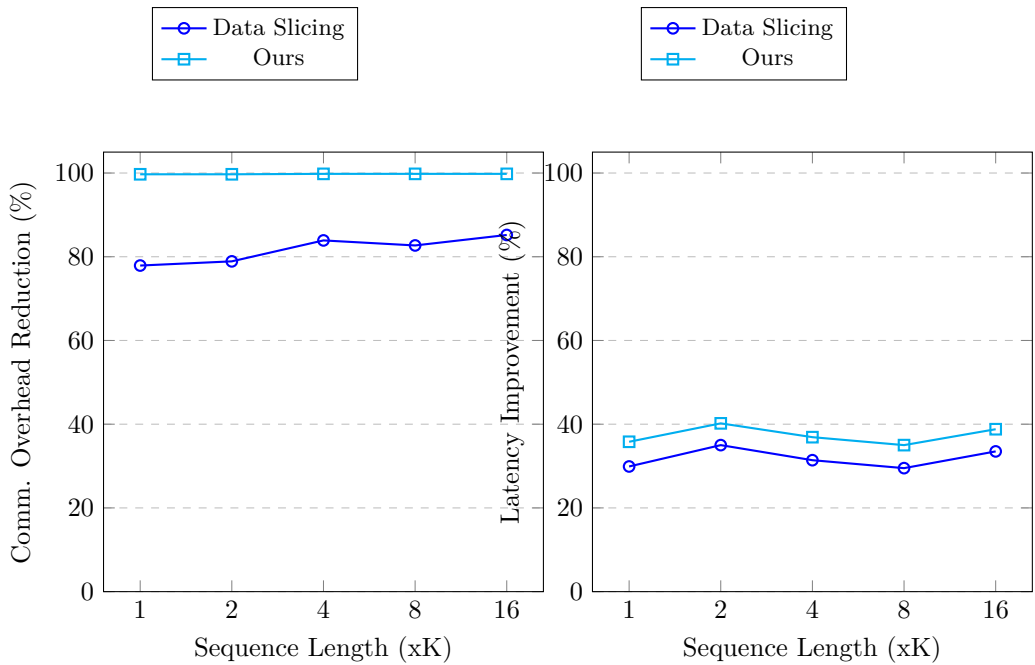

\paragraph{Scaling with Input Size.}
We compare the performance of our method and the data slicing approach against the vanilla baseline under varying input sizes. Figure~\ref{fig:comm_reduction_scaling} illustrates the communication overhead reduction (higher is better) relative to the baseline. Our method consistently hides over 99\% of the communication overhead across all sequence lengths, demonstrating robust scalability.

The experimental configuration uses $d_{\text{model}}=4096$ and tensor parallelism (TP)$=4$. For input scaling, we start with a sequence length of 1K and batch size of 128, and progressively double the sequence length while halving the batch size to maintain memory constraints. This results in input shapes $(b,s) = (128,1024)$, $(64,2048)$, $(32,4096)$, $(16,8192)$, and $(8,16384)$. The latency observed for these configurations for baseline, data slicing and our method are presented in Table~\ref{tab:mlp_scaling_latency}.

\begin{table}[h!]
\caption{Example resized table with fixed column widths}
\label{tab:mlp_scaling_latency}
\centering
    \resizebox{\textwidth}{!}{
        \begin{tabular}{
            |p{2.6cm}
            |p{2.5cm}
            |p{2.5cm}
            |p{2.5cm}
            |p{2.5cm}
            |p{2.5cm}|
            }
            \hline
            \multirow{2}{*}{Method} & \multicolumn{5}{c|}{Sequence Length} \\
            \cline{2-6}
             & 1K & 2K & 4K & 8K & 16K \\
            \hline
             Slicing latency (ms) & 83.8 & 83.7 & 83.9 & 83.7 & 83.9 \\
            \hline
             Ours Latency (ms) & 77.1 & 77.2 & 77.1 & 77.2 & 77.2  \\
            \hline
            \hline
             Improvement (\%) ($\uparrow$)& 8.0 & 7.9 & 8.1 & 7.8 & 8.0 \\
            \hline
        \end{tabular}
    }
\end{table}

Table~\ref{tab:scaling_combined} summarizes both the communication overhead reduction and end-to-end latency improvement relative to the vanilla baseline. The data slicing approach achieves communication reduction in the range of 77.9\% to 85.2\%, whereas our method consistently achieves approximately 99.7\%--99.8\% reduction across all sequence lengths. In terms of latency, our method also consistently outperforms the data slicing approach, delivering higher end-to-end latency improvements for all configurations. These results demonstrate that the proposed method not only effectively hides communication overhead but also translates this advantage into tangible end-to-end performance gains across a wide range of input scales. Figure~\ref{fig:comm_reduction_scaling} summarizes the trend over sequence length showing consistent improvement.

\begin{table}[t]
\centering
\caption{Communication overhead reduction (\%) and end-to-end latency improvement (\%) compared to the vanilla baseline across sequence lengths.}
\label{tab:scaling_combined}
\resizebox{0.75\linewidth}{!}{
\begin{tabular}{l l c c c c c}
\toprule
\multirow{2}{*}{\textbf{Experiment}} & \multirow{2}{*}{\textbf{Method}} & \multicolumn{5}{c}{\textbf{Sequence Length}} \\
\cmidrule(lr){3-7}
 & & \textbf{1K} & \textbf{2K} & \textbf{4K} & \textbf{8K} & \textbf{16K} \\
\midrule
\multirow{2}{*}{Comm. Overhead Reduction (\%)} & Data Slicing & 77.9 & 78.9 & 83.9 & 82.7 & 85.2 \\
 & Ours         & 99.7 & 99.7 & 99.8 & 99.8 & 99.8 \\
\midrule
\multirow{2}{*}{Latency Improvement (\%)} & Data Slicing & 29.9 & 35.0 & 31.4 & 29.5 & 33.5 \\
 & Ours         & 35.8 & 40.2 & 36.9 & 35.0 & 38.8 \\
\bottomrule
\end{tabular}
}
\end{table}

\paragraph{Attention Layer.}
For the attention layer, we consider two instantiation of CommFuse-TPSP. The first one uses the newly proposed row parallel linear projection with the FuseRS as a drop in replacement of traditional chunking based row parallel linear projection used in Megatron. The second implementation is for overlaping more compute using the proposed query split attention with FuseRS. We compare both of the implementation with a baseline without overlap and conventional chunking/data decomposing based approach. The comparison of latency for the forward pass over an attention layer for different data shapes are reported in table~\ref{tab:latency_attn}.

\begin{table}
\centering
\resizebox{0.7\columnwidth}{!}{
\begin{tabular}{|c|c|c|c|c|}

\hline
\multirow{2}{*}{Data (B x S) }  & \multicolumn{4}{c|}{Latency (ms) (Change $\%$ ) }  \\ 
\cline{2-5} 
                  & Baseline & Slicing & Row Parallel & Query Split  \\ \hline
               $64 \times 1024 $   & 196.2 & 179.6 & 170.7 & 164.2 \\ \hline
               $32 \times 2048 $   & 222.8 & 202.2 & 200.7 & 181.2 \\ \hline
               $16 \times 4096 $   & 270.1 & 246.1 & 230.9 & 215.8 \\ \hline
               $4  \times 8192 $   & 195.6 & 279.7 & 174.6 & 145.1 \\ \hline
 
\end{tabular}
}
\caption{Latency for the forward pass of an \emph{attention} layer for different sequence lengths. Here B and S are batch and sequence length; all data uses same embedding dimension of 4096.}
\label{tab:latency_attn}
\end{table}

\section{Conclusion}
The method presented in this work offers an exact and broadly applicable solution for achieving full compute–communication overlap in intra‑layer model parallelism across modern deep neural network architectures. By eliminating approximation and preserving model accuracy, it provides a practical path to improving latency and throughput in large‑scale training and inference, particularly for transformer‑based LLMs, vision transformers, Mamba models, and multimodal systems. The approach generalizes across attention, MLP, and Mamba layers, supports both TPSP and UP parallelism, and applies consistently from pretraining through inference. As model sizes and deployment demands continue to grow, the ability to overlap computation with communication at fine granularity offers a scalable and architecture‑agnostic mechanism for improving efficiency in contemporary AI workloads.

\subsubsection*{Broader Impact Statement}
This work introduces a method for improving the efficiency of parallelism in large‑scale neural networks. By enabling full compute–communication overlap across transformer, Mamba, and hybrid architectures, the method can reduce latency during both training and inference without affecting model accuracy. These efficiency gains may help lower the resource requirements associated with large language models, vision transformers, and multi-modal systems, potentially reducing energy consumption and broadening access to high‑capacity models in research and industry.

At the same time, increased efficiency may accelerate the deployment of large models in real‑world applications, amplifying both their benefits and their risks. More accessible large‑scale models can support advances in education, accessibility, scientific discovery, and automation. On the other hand, they may also exacerbate concerns related to fairness, misuse, and the societal impact of widespread AI adoption. As with any enabling technology, the broader consequences depend on how the resulting systems are developed, governed, and applied. We encourage practitioners to pair the technical improvements presented here with responsible deployment practices and ongoing evaluation of downstream impacts.


\bibliography{tmlr}
\bibliographystyle{tmlr}

\appendix

\section{Appendix: Illustrative Example of FuseRS Decomposition}
\label{sec:fuse_rs_example}

\subsection{Overview}

Our solution belongs to the class of methods that decompose collective communication into peer-to-peer (P2P) communications to enable overlap with computation. We propose multiple decomposition strategies that are generally applicable across models.


\subsection{Running Example}
Here we present an example case for FuseRS. Consider a setup with four devices: $A, B, C, D$. Each device initially holds a sequence:
\begin{align*}
A &: [a_1, a_2, a_3, a_4] \\
B &: [b_1, b_2, b_3, b_4] \\
C &: [c_1, c_2, c_3, c_4] \\
D &: [d_1, d_2, d_3, d_4]
\end{align*}

After a reduce-scatter operation, the result is:
\begin{align*}
[a_1 + b_1 + c_1 + d_1], \;
[a_2 + b_2 + c_2 + d_2], \;
[a_3 + b_3 + c_3 + d_3], \;
[a_4 + b_4 + c_4 + d_4]
\end{align*}

To achieve this, each device must send data slices to others:
\begin{itemize}
    \item $A \rightarrow \{B, C, D\}$: $a_2, a_3, a_4$
    \item $B \rightarrow \{C, D, A\}$: $b_3, b_4, b_1$
    \item $C \rightarrow \{D, A, B\}$: $c_4, c_1, c_2$
    \item $D \rightarrow \{A, B, C\}$: $d_1, d_2, d_3$
\end{itemize}

Here are the various ways of scheduling decomposed communications, which we used in empirical ablation.

\subsection{(a) Pairwise Bidirectional P2P}

The $12$ send operations can be decomposed into $6$ bidirectional peer-to-peer (P2P) exchanges across $3$ rounds:

\begin{itemize}
    \item Round 1: $(A \leftrightarrow B), (C \leftrightarrow D)$
    \item Round 2: $(A \leftrightarrow C), (B \leftrightarrow D)$
    \item Round 3: $(A \leftrightarrow D), (B \leftrightarrow C)$
\end{itemize}

This schedule ensures:
\begin{itemize}
    \item $n/2$ simultaneous transfers per round,
    \item Completion in $n-1$ rounds (for even $n$),
    \item Elimination of tail overhead by overlapping communication with computation.
\end{itemize}

This pattern is analogous to a round-robin tournament where each participant plays every other participant exactly once.

\subsection{(b) Circular P2P}

An alternative approach is a unidirectional ring communication pattern:
\[
A \rightarrow B \rightarrow C \rightarrow D \rightarrow A
\]

Each device maintains a buffer ($m_1, m_2, m_3, m_4$), initialized to zero.

\paragraph{Round-wise Execution}
\begin{itemize}
    \item \textbf{Round 1:} \\
    $m_1 = b_1,\; m_2 = c_2,\; m_3 = d_3,\; m_4 = a_4$ (rotated clockwise)

    \item \textbf{Round 2:} \\
    $m_1 = b_1 + c_1,\; m_2 = c_2 + d_2,\; m_3 = d_3 + a_3,\; m_4 = a_4 + b_4$

    \item \textbf{Round 3:} \\
    $m_1 = b_1 + c_1 + d_1,\; m_2 = c_2 + d_2 + a_2,\; m_3 = d_3 + a_3 + b_3,\; m_4 = a_4 + b_4 + c_4$

    \item \textbf{Round 4:} \\
    $m_1 = a_1 + b_1 + c_1 + d_1$, \\
    $m_2 = a_2 + b_2 + c_2 + d_2$, \\
    $m_3 = a_3 + b_3 + c_3 + d_3$, \\
    $m_4 = a_4 + b_4 + c_4 + d_4$
\end{itemize}

At the end of Round 4, each device holds its final reduced value. No further communication is required.

This method:
\begin{itemize}
    \item Uses strictly local, circular communication,
    \item Avoids tail latency,
    \item Naturally overlaps communication and computation.
\end{itemize}

\subsection{(c) Circular Computation Slices}

A third approach reorganizes computation itself in a circular fashion over data slices. This improves memory locality by ensuring contiguous access across rounds.

In this scheme:
\begin{itemize}
    \item Communication alternates between circular ring transfers and pairwise P2P,
    \item Computation is aligned with data movement,
    \item Subsequent rounds benefit from improved cache and memory efficiency.
\end{itemize}


\section{Appendix: Illustrative Example of Query Split Attention}
\label{sec:query_split_attention_example}
\begin{figure}[h!]
\centering

\begin{subfigure}{\textwidth}
\centering
\resizebox{0.8\textwidth}{!}{
    \includegraphics{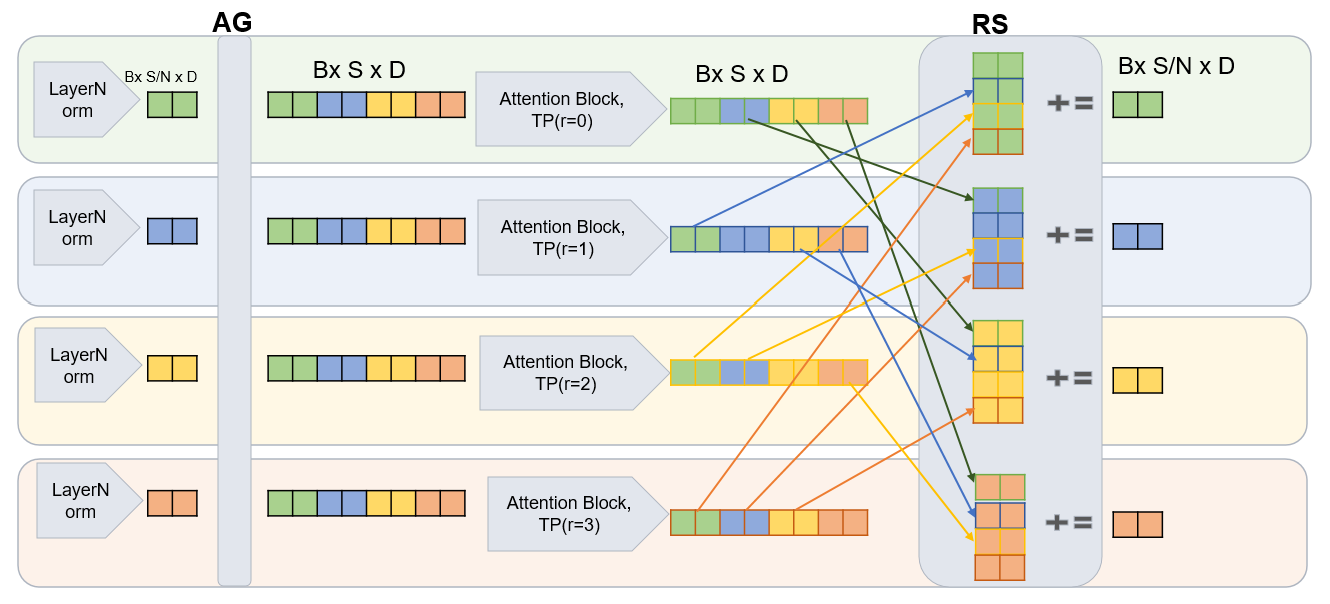}
}
\caption{Attention with tensor parallelism (TP) using non overlapped reduce-scatter resulting in communication overhead.}
\label{fig:subfig1}
\end{subfigure}

\vspace{0.5cm}

\begin{subfigure}{\textwidth}
\centering
\resizebox{0.8\textwidth}{!}{
    \includegraphics{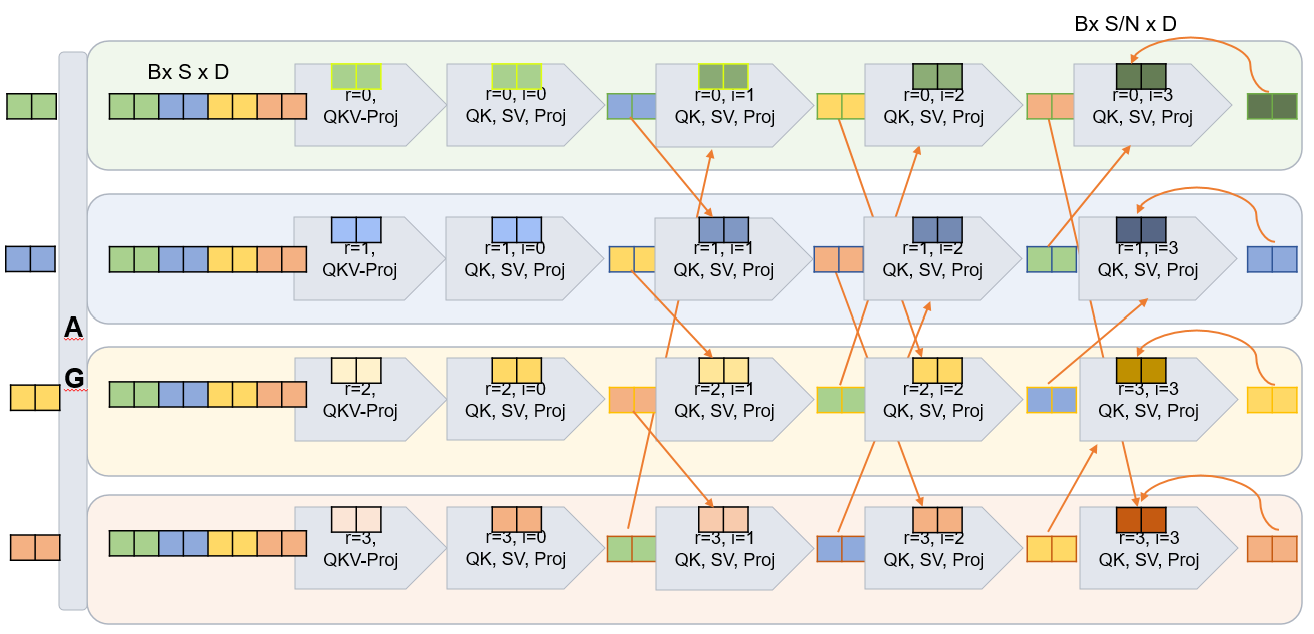}
}
\caption{Computation of the Query-Split FuseRS Attention with TP using asynchronous P2P decomposed communication overlapping with computations. }
\label{fig:subfig2}
\end{subfigure}

\caption{Example of decomposed query split attention for FuseRS showing the benefit of asynchronous P2P decomposed communication eliminating communication overhead.}
\label{fig:visual4qsattn}

\end{figure}

Figure~\ref{fig:visual4qsattn} presents a visual comparison of the computation and communication patterns in Tensor Parallel attention using reduce-scatter and those of the proposed Query-Split attention..

\end{document}